\pdfoutput=1

\documentclass[11pt]{article}

\usepackage[final]{acl2023}

\usepackage{times}
\usepackage{latexsym}
\usepackage{booktabs}
\usepackage{CJKutf8}
\usepackage[T1]{fontenc}

\usepackage[utf8]{inputenc}

\usepackage{microtype}
\usepackage{latexsym}
\usepackage{inconsolata}

\usepackage{graphicx}  
\usepackage{float}  
\usepackage{subfigure}  

\title{USTCCTSU at SemEval-2024 Task 1: Reducing Anisotropy for Cross-lingual Semantic Textual Relatedness}

\author{Jianjian Li\textsuperscript{\rm 1 \thanks{ \quad Equal contribution and shared co-first authorship.} } \quad
        Shengwei Liang\textsuperscript{\rm 1 \footnotemark[1] } \quad
        Yong Liao\textsuperscript{1,2 $\dagger$} \\
        \textbf{Hongping Deng\textsuperscript{2}} \quad
        \textbf{Haiyang Yu\textsuperscript{2}}\thanks{\quad Corresponding author.} \\
1. University of Science and Technology of China, CCCD Key Lab of MCT \\
2. Institute of Dataspace \\
\texttt{\{sa22221088,sewell\}@mail.ustc.edu.cn}
        }

\begin{document}
\maketitle

\begin{abstract}

Cross-lingual semantic textual relatedness task is an important research task that addresses challenges in cross-lingual communication and text understanding. It helps establish semantic connections between different languages, crucial for downstream tasks like machine translation, multilingual information retrieval, and cross-lingual text understanding.
Based on extensive comparative experiments, we choose the $XLM$-$R_{base}$ as our base model and use pre-trained sentence representations based on whitening to reduce anisotropy.
Additionally, for the given training data, we design a delicate data filtering method to alleviate the curse of multilingualism.
With our approach, we achieve a \textbf{2nd} score in Spanish, a \textbf{3rd} in Indonesian, and multiple entries in the top ten results in the competition's track C.
We further do a comprehensive analysis to inspire future research aimed at improving performance on cross-lingual tasks.

\end{abstract}

\section{Introduction}






Semantic textual relatedness (STR) encompasses a broader concept that takes into account various commonalities between two sentences. This includes factors such as being on the same topic, expressing the same viewpoint, originating from the same period, one sentence elaborating on or following from the other, and more. SemEval is an international workshop on semantic evaluation. In track C of SemEval-2024 task 1: Cross-lingual \cite{semrel2024task}, participants are to submit systems,which are developed without the use of any labeled semantic similarity or semantic relatedness datasets in the target language and with the use of labeled datasets \cite{semrel2024dataset} from at least one other language.

Various methods were proposed to address the task of textual relatedness. One common approach is based on feature engineering, where the syntactic, semantic, and structural features of text, such as word frequency, TF-IDF, and word embeddings, are extracted. Machine learning algorithms are then employed for relatedness determination. Another popular approach is based on deep learning methods, such as Convolutional Neural Networks \cite{lecun1989backpropagation}, Recurrent Neural Networks \cite{graves2012long} , and self-attention mechanisms \cite{vaswani2017attention}. These methods can capture semantic relationships and contextual information within the text, and they are trained on large-scale datasets to enhance model performance and generalization ability.
\begin{figure}[t]
    \centering
    \includegraphics[width=\columnwidth]{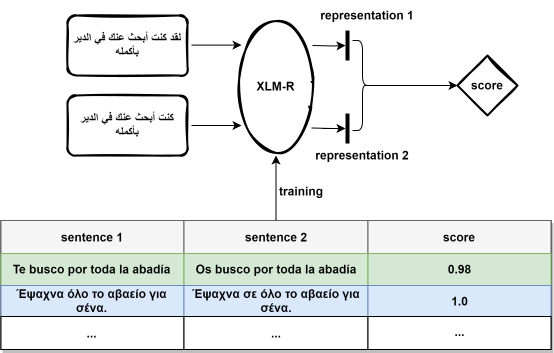}
    \caption{The description of cross-lingual semantic textual relatedness task.} \label{figure:taskdescription}
\end{figure}

However, there are two challenges in track C of SemEval-2024 task 1:
\begin{itemize}
    \item Compared with static word representation such as Word2Vec \cite{mikolov2013efficient} and Glove \cite{pennington2014glove}, the pretrained language models (PLM) can obtain sentence representation for different sentence in different contexts, thereby solving different problems. However, the vectors of BERT-based PLM models have limitations: \textbf{BERT-based models always induces a non-smooth anisotropic semantic space of sentences, which harms its performance of semantic similarity} \cite{gao2019representation,li2020sentence}, which can lead to a challenge that sentences are strikingly similar while using the cosine similarity metric.  
    \item Participants are not allowed to utilize labeled datasets in the target language for training. Instead, they must use labeled data in different languages as the training set to train the model and provide predictions in the target language. However, \textbf{multilingual pre-trained models suffer from the curse of multilingualism} \cite{conneau2020unsupervised}, that is, the overall performance of both monolingual and cross-lingual baselines declines when adding more languages to training data over a certain point. Hence, it is essential to investigate which additional languages would be  inefficient as the training dataset for the target language.
\end{itemize}
In this paper, we used whitening techniques \cite{su2021whitening}, which maps vectors to standard orthogonal bases, to transform the word vector representations from anisotropic to isotropic, and surprisingly, we found that whitening significantly improves the accuracy of judging semantic similarity. Given the absence of labeled data in the target language, it is difficult to determine which other language would yield better prediction results when used as training data. Therefore, we proposed that removing certain language categories from the training data for a specific target language contributed to improving performance.


We conducted extensive experiments to demonstrate the effectiveness of the method we employed. As a result, our submitted outcomes achieved a \textbf{2nd} score in Spanish and a \textbf{3rd} score in Indonesian in track C of SemEval-2024's task 1. Additionally, we obtained multiple top-ten rankings in the competition.
\section{Background}

The task of semantic text relatedness covers several specific subtasks, including semantic similarity, semantic matching, textual entailment, semantic relation classification, and text pair ranking. Previous work has proposed various methods for these specific tasks, such as: Lexical and syntactic-based methods \cite{gamallo2001syntactic,pakray2011textual}: These methods rely on lexical and syntactic rules, such as word vector matching, lexical overlap, and syntactic tree matching. However, these methods often fail to capture higher-level semantic relationships.
Feature engineering-based machine learning methods \cite{chia2021machine,fan2019deep}: These methods involve using manually designed features, such as bag-of-words models \cite{zhang2010understanding}, tf-idf weights, and syntactic features, followed by using machine learning algorithms like support vector machines and random forests for prediction.

While these methods have improved performance to some extent, they still have limitations in capturing complex semantic relationships.
Neural network-based models: These models use neural networks to learn representations of text and capture semantic relationships between texts through training data. This includes methods that fine-tune pre-trained language models (e.g., BERT \cite{kenton2019bert} and GPT2 \cite{radford2019language} etc.), as well as approaches that employ Siamese networks, LSTM, CNN, and other architectures for text encoding and matching.
Transfer learning and multi-task learning \cite{pilault2020conditionally,wu2020understanding}: These methods leverage knowledge from pre-trained models on related tasks to improve the performance of semantic textual relatedness tasks through transfer learning \cite{koroleva2019measuring}. Multi-task learning combines multiple related tasks in training to enhance the model's generalization ability and effectiveness.
Application of external knowledge resources: Researchers have also attempted to incorporate external knowledge resources such as word embeddings, semantic knowledge graphs, and multilingual data to enhance the model's understanding of semantic relationships.

For cross-lingual semantic similarity tasks, mapping texts from different languages into a shared semantic space for similarity calculation is necessary. To address this, researchers have proposed various cross-lingual representation learning methods. Among them, unsupervised alignment methods like unsupervised machine translation \cite{lample2017unsupervised} and cross-lingual pre-training models \cite{liang2020xglue} can learn the correspondences between multiple languages and map texts to a shared vector space.

However, \citep{conneau2019cross} and \citep{wangimproving} mentioned that vector representations based on the Transformer models  exhibit anisotropy, which means that the vectors are unevenly distributed and clustered in a narrow cone-shaped space. Therefore, both Bert-flow \cite{li2020sentence} and Bert-whitening \cite{su2021whitening} aim to address the same issue, which is the anisotropy and uneven distribution of sentence embeddings.




\section{System Overview}

\subsection{Framework Overview}


In this section, we will introduce our proposed method for STR task which has three main modules.

\begin{itemize}
\item \textbf{PLM Encoder}
We adopted the pretrained language model XLM-RoBERTa-base ($XLM$-$R_{base}$) \cite{conneau2020unsupervised} for initial sentence encoding, which combines two powerful models: Transformer and RoBERTa. $XLM$-$R_{base}$ demonstrates strong multilingual capabilities and a deep understanding of semantics, surpassing some monolingual pre-training models. After conducting a series of tests on mBERT \cite{pires2019multilingual}, XLM \cite{conneau2019cross}, and $XLM$-$R_{base/large}$, we selected $XLM$-$R_{base}$ as the encoder due to its superior performance.


\item \textbf{Whitening Module}
After obtaining the sentence vectors of two utterances using $XLM$-$R_{base}$, we could have directly calculated the cosine similarity between the two vectors, but the sentence vectors after $XLM$-$R_{base}$ show anisotropy between them and are distributed in a conical space, resulting in a high cosine similarity. Therefore, we introduce the Whitening module to change the distribution of the sentence vector space so that its distribution has various anisotropies, amplifying the differences between the vectors and stimulating the performance of $XLM$-$R_{base}$ on the semantic text similarity reading task.

\item \textbf{Data Filtering}
The authors of \cite{conneau2020unsupervised} mention the curse of multilingualism, where adding more languages leads to an improvement in cross-lingual performance for low-resource languages up to a certain point, after which the overall performance of both monolingual and cross-lingual baselines declines. In the task of cross-lingual semantic text similarity, to maximize the exploration of the positive impact of other languages on the target language, we propose a new dataset selection method. As the influence between languages is mutual, we utilize the unlabeled data of the target language to detect the impact of each language in track A, excluding the target language, and infer its influence on the target language. This allows us to select the training dataset optimally. This approach helps eliminate interference from certain languages on the target language and avoids the curse of multilingualism.
\end{itemize}

\subsection{PLM Encoder}

Through a simple test and comparative analysis of different multilingual pre-training models, we found that $XLM$-$R_{base}$ outperforms mBERT. $XLM$-$R_{base}$ is a cross-lingual pre-training model based on the BERT architecture, an improvement and extension of the original XLM model. The goal of $XLM$-$R_{base}$ is to enhance the performance and effectiveness of multilingual text processing. $XLM$-$R_{base}$ utilizes larger-scale pre-training data and more sophisticated training methods to enhance the model's representation capabilities. It undergoes deep learning on a large amount of unsupervised data using RoBERT \cite{liu2019roberta} technology. This enables $XLM$-$R_{base}$ to better understand and capture the semantic and grammatical features between different languages. Compared to the original XLM, $XLM$-$R_{base}$ has made several improvements. Firstly, it introduces a dynamic masking mechanism that allows the model to better perceive contextual information. Secondly, $XLM$-$R_{base}$ emphasizes cross-lingual consistency learning through adversarial training, enabling better alignment and sharing of model parameters. This enables $XLM$-$R_{base}$ to provide more accurate representations of texts in cross-lingual tasks. Compared to mBERT, $XLM$-$R_{base}$ employs larger-scale pre-training data, covers more languages, and incorporates improvements through RoBERTa technology. This enables $XLM$-$R_{base}$ to better learn and capture the semantic and grammatical features between different languages, thereby enhancing the model's representation capabilities and performance.

\subsection{Whitening Module}
Due to the existence of anisotropy among the vectors obtained from the initial encoding by $XLM$-$R_{base}$, cosine similarity cannot accurately measure the semantic similarity between sentences. Therefore, we chose to use whitening to map the original vector space to an isotropic space, where the vectors are transformed into vectors in a standard orthogonal bases. 
The principle is as follows:

Suppose we have a set of sentence vectors $ S=\{s_1,s_2, \ldots,s_n \} $, the set of vectors can be transformed into a set of vectors with isotropy (i.e., zero mean and a covariance matrix of the identity matrix) through the following transformation $ \tilde{S}=\{\tilde{s}_1,\tilde{s}_2,\ldots, \tilde{s}_n \} $.
\begin{equation}
    \tilde{s}_i = (x_i - \mu)\mathbf{W} \label{1}
\end{equation}  
If we want to make the set $\tilde{S} $ have a zero mean, we need to:
\begin{equation}
    \mu = \frac{1}{n} \sum\limits_{i=1}^n {s_i} \label{2}
\end{equation}
The next step is to calculate $\mathbf{W}$. The covariance matrix of $S$:
\begin{equation}
    \Sigma = \frac{1}{n} \sum\limits_{i=1}^{n} {\mathbf{(s_i - \mu)}^\top(s_i - \mu)} \label{3}
\end{equation}
The covariance matrix of $\tilde{S}$:
\begin{equation}
    \tilde{\Sigma}= \mathbf{W}^\top \Sigma \mathbf{W} \label{4}
\end{equation}
If we want to transform $\tilde{\Sigma}$ into the identity matrix $I$, we need to:
\begin{equation}
    \tilde{\Sigma}= \mathbf{W}^\top \Sigma \mathbf{W} =\mathbf{I}\label{5}
\end{equation}
Then:
\begin{equation}
    \Sigma = (\mathbf{W}^\top)^{-1} \mathbf{W}^{-1} = \mathbf{(W^{-1})}^\top \mathbf{W}^{-1}  \label{6}
\end{equation}
Since $\Sigma$ is a positive definite symmetric matrix as the covariance matrix, it can be decomposed using Singular Value Decomposition (SVD), yielding:
\begin{equation}
    \Sigma = \mathbf{U} \Lambda \mathbf{U}^\top \label{7}
\end{equation}
By combining equations (6) and (7), we obtain:
\begin{equation}
    \mathbf{(W^{-1})}^\top \mathbf{W}^{-1}=\mathbf{U} \Lambda \mathbf{U}^\top = \mathbf{U} \sqrt{\Lambda} \sqrt{\Lambda} \mathbf{U}^\top  \label{8}
\end{equation}
Then:
\begin{equation}
    \mathbf{(W^{-1})}^\top \mathbf{W}^{-1}= (\sqrt{\Lambda}\mathbf{U}^\top)^\top \sqrt{\Lambda} \mathbf{U}^\top  \label{9}
\end{equation}
Therefore, we can obtain $\mathbf{W^{-1}}=\sqrt{\Lambda}\mathbf{U}$, and finally obtain $\mathbf{W}$ as follows:
\begin{equation}
    \mathbf{W}=\mathbf{U}\sqrt{\Lambda^{-1}}
\end{equation}

\subsection{Data Filtering}
Our experiments have shown that when selecting training data for the target language, using a mixture of multiple languages often yields better results than using a single language. The authors of the $XLM$-$R_{base}$ paper mentioned that incorporating more languages improves the cross-lingual performance of low-resource languages up to a certain point. Beyond that point, the overall performance of both monolingual and cross-lingual benchmarks starts to decline. Additionally, we believe that there is interdependence between languages. For example, if including text from language A in training set to compute whitening parameters leads to a decrease in the prediction performance for language B, we expect that the opposite would hold true as well.

Therefore, inspired by this insight, we used the text in the target language as the dataset and individually tested the labeled training data provided in track A for different languages.
For example, if the target language is identified  by $T$, we use the text of $T$ for whitening, and test the performance on language ${Test_A}$, ${Test_B}$, ${Test_C}$, ${Test_D}$,... one by one.
If the prediction performance of ${Test_A}$ decreases after using  $T$ compared to not using  $T$ (measured by the Spearman correlation \cite{myers2004spearman} between the gold labels and predicted labels obtained using language ${Test_A}$), then ${Test_A}$ is excluded from target language's training set.

In the case of the Spanish, using the training set without data filtering (1,000 each of all data except Spanish) resulted in a final spearman coefficient of 0.6375; using the training set with data filtering (1000 each of kin and ind) resulted in a final spearman coefficient of 0.6886. Although the training data for about ten languages were reduced, the results were are significantly improved.


\section{Experimental Setup}

\begin{figure*}[t]
    \centering
    \subfigbottomskip = 2pt 
	\subfigcapskip = -5pt 
	\subfigure[without whitening]{
		\includegraphics[width=0.48\linewidth]{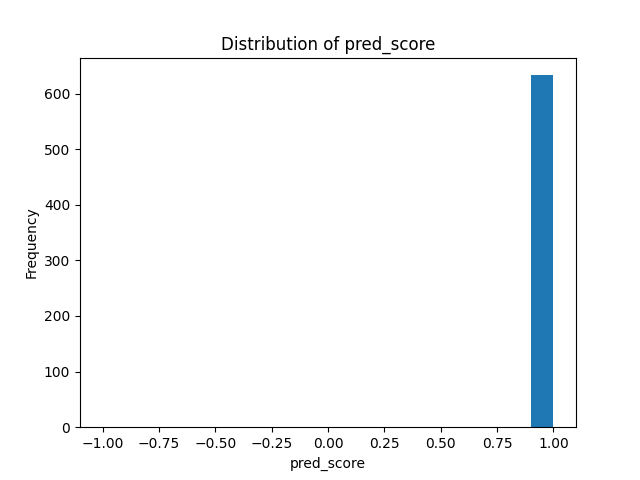}}
	\subfigure[with whitening]{
		\includegraphics[width=0.48\linewidth]{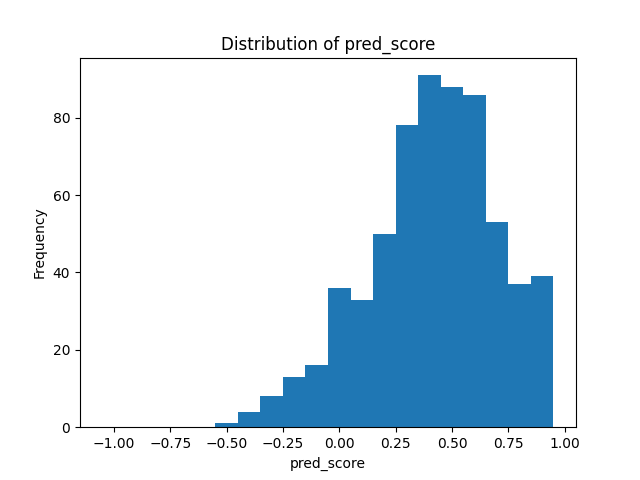}}
    \caption{The results of model without whitening and with whitening.} \label{figure:resultdistribution}
\end{figure*}

We use the 12 labeled training data from \citep{semrel2024dataset} as training data and the test data from track C as test data. We observe that the amount of data for each language is concentrated around 1,000, so we take 1,000 as the boundary, use oversampling to make up for less than 1,000, and use randomization to take out 1,000 for more than 1,000 to ensure that sentence pairs of different similarities are involved. In finding the training set combinations for the target languages, we compute the $\mu$ and $\mathbf{W}$ of whitening using the text data of the target languages in track C. We predict the training data one by one for each language, and compute the spearman coefficients using the gold labels and the predicted labels of the training data, and compare the results with the data without any whitening (i.e., the prediction result of the base model) to evaluate whether the target language enhances a certain language in the training data or not, and if it does not, it is excluded from the train data. Eventually, the remaining language data is used as a training set to predict the target language.

The hyperparameters are set as follows: we choose to freeze the pretrained model $XLM$-$R_{base}$ while setting the top\-k parameter of whitening to 256.
The rubric we used was the spearman coefficient, calculated using the methodology provided by the competition officials.
	

\section{Results}

The official competition used the spearman coefficients to evaluate the results, and Table 1 gives the results of the spearman coefficients for both Indonesian (ind) and Spanish (esp) languages throughout the experiment. There is a big difference in the multilingual ability of different model bases. We chose $XLM$-$R_{base}$, which performs better, and we can see that the overall results are improved after using the whitening module to transform the vector space; $XLM$-$R_{base}$ with whitening is better than baseline, and we got a good ranking in track C of SemEval-2024 task 1, in which we ranked second in esp and third in ind.

\begin{table}[ht]
    \begin{tabular}{l c c}
    \hline
        \textbf{} & \textbf{ind-test} & \textbf{esp-test} \\ \hline
        Baseline & 0.4700 & 0.6200 \\ \hline
        mBERT & 0.4390 & 0.5971 \\ 
        $XLM$-$R_{base}$ & 0.4390 & 0.5907 \\ 
        $XLM$-$R_{large}$ & 0.4267 & 0.6003 \\ \hline
        mBERT-whitening & 0.4471 & 0.6411 \\ 
        $XLM$-$R_{base}$-whitening & 0.4746 & \textbf{0.6886} \\ 
        $XLM$-$R_{large}$-whitening & \textbf{0.4845} & 0.6648 \\ \hline
    \end{tabular}
    \caption{The spearman coefficient of different models and baseline.}
\end{table}
As can be seen from Table 1, the whitening module improves the STR task more significantly,the baseline is given by \cite{semrel2024dataset}. In order to further verify whether whitening works, we counted the cosine similarity distribution statistics of the data without whitening processing and after whitening. Figure~\ref{figure:resultdistribution} gives two cosine similarity statistics. The left side is the cosine similarity statistics without whitening. The cosine similarity of all utterance pairs is concentrated between 0.9 and 1.0, indicating that the vector space is anisotropic. In contrast, after adding whitening, the whole distribution tends to be normal, which indicates that whitening plays a role in mapping the vectors to an isotropic space, amplifying the differences between statements.


	
	

\section{Conclusion}

We use $XLM$-$R_{base}$ with whitening and propose a dataset filtering method that exploits the positive correlation of linguistic interactions, achieving good rankings in SemEval-2024 task 1 track C. We verifies that whitening performs well on utterance characterization as well as STR task. Besides, the proposed dataset filtering method is more efficient and can alleviate the multilingual curse problem in cross-language problems to some extent.

In the future, we will further study this positive correlation of language interactions, and we hope that this correlation can become more detailed, not only in terms of inter-language correlations but also in terms of the domain of the text. We also hope that this correlation can be better utilized in dataset preprocessing, not only to eliminate poorly performing languages but to further improve the combination of datasets that can be directly selected to correspond to the optimal solution.

\section*{Acknowledgments}

We  want to express gratitude to the anonymous reviewers for their hard work and kind comments, which will further improve our work in the future.
This work is funded by national key research and development program under grant 2021YFC3300500-02.

\bibliography{custom}

\appendix



\end{document}